\documentclass[10pt,twocolumn,letterpaper]{article}

\usepackage{cvpr}
\usepackage{times}
\usepackage{epsfig}
\usepackage{graphicx}
\usepackage{amsmath}
\usepackage{amssymb}
\usepackage{subcaption}
\usepackage{caption}
\usepackage{pifont}
\usepackage{multirow}
\usepackage{boldline}

\usepackage{float}
\usepackage{pdfpages}
\usepackage{subcaption}
\newcommand{\RNum}[1]{\uppercase\expandafter{\romannumeral #1\relax}}
\newcommand{\tabincell}[2]{\begin{tabular}{@{}#1@{}}#2\end{tabular}}
\DeclareMathAlphabet      {\mathbfit}{OML}{cmm}{b}{it}

\usepackage[breaklinks=true,bookmarks=false]{hyperref}

\cvprfinalcopy % *** Uncomment this line for the final submission

\def\cvprPaperID{****} % *** Enter the CVPR Paper ID here

\ifcvprfinal\fi
\begin{document}

%%%%%%%%% TITLE
\title{Growing a Brain: Fine-Tuning by Increasing Model Capacity}

\author{Yu-Xiong Wang\qquad Deva Ramanan\qquad Martial Hebert\\
Robotics Institute, Carnegie Mellon University\\
{\tt\small \{yuxiongw,dramanan,hebert\}@cs.cmu.edu}
}

\maketitle
%\thispagestyle{empty}

%%%%%%%%% ABSTRACT
\begin{abstract}
CNNs have made an undeniable impact on computer vision through the ability to learn high-capacity models with large annotated training sets. One of their remarkable properties is the ability to transfer knowledge from a large source dataset to a (typically smaller) target dataset.  This is usually accomplished through fine-tuning a fixed-size network on new target data. Indeed, virtually every contemporary visual recognition system makes use of fine-tuning to transfer knowledge from ImageNet. In this work, we analyze what components and parameters change during fine-tuning, and discover that increasing model capacity allows for more natural model adaptation through fine-tuning. By making an analogy to developmental learning, we demonstrate that ``growing'' a CNN with additional units, either by widening existing layers or deepening the overall network, significantly outperforms classic fine-tuning approaches. But in order to properly grow a network, we show that newly-added units must be appropriately normalized to allow for a pace of learning that is consistent with existing units. We empirically validate our approach on several benchmark datasets, producing state-of-the-art results.
\end{abstract}

%%%%%%%%% BODY TEXT
\section{Motivation}
Deep convolutional neural networks (CNNs) have revolutionized visual understanding, through the ability to learn ``big models'' (with hundreds of millions of parameters) with ``big data'' (very large number of images). Importantly, such data must be annotated with human-provided labels. Producing such massively annotated training data for new categories or tasks of interest is typically unrealistic. Fortunately, when trained on a large enough, diverse ``base'' set of data (\eg, ImageNet), CNN features appear to  transfer across a broad range of tasks~\cite{razavian2014cnn,azizpour2015generic,yosinski2014transferable}. However, an open question is how to best adapt a pre-trained CNN for novel categories/tasks. 

{\bf Fine-tuning:} Fine-tuning is by far the dominant strategy for transfer learning with neural networks~\cite{oquab2014learning,azizpour2015generic,razavian2014cnn,yang2015multi,girshick2014rich,hariharan2015hypercolumns}. This approach was pioneered in~\cite{hinton2006reducing} by transferring knowledge from a generative to a discriminative model, and has since been generalized with great success~\cite{girshick2014rich,zeiler2014visualizing}. The basic pipeline involves replacing the last ``classifier'' layer of a pre-trained network with a new randomly initialized layer for the target task of interest. The modified network is then fine-tuned with additional passes of appropriately tuned gradient descent on the target training set. Virtually {\em every} contemporary visual recognition system uses this pipeline. Even though its use is widespread, fine-tuning is still relatively poorly understood. For example, what fraction of the pre-trained weights actually change and how? 

\begin{figure}[t]
\begin{center}
\centering
   \includegraphics[trim=1cm 3.7cm 1.8cm 2.3cm,clip=true,width=\linewidth]{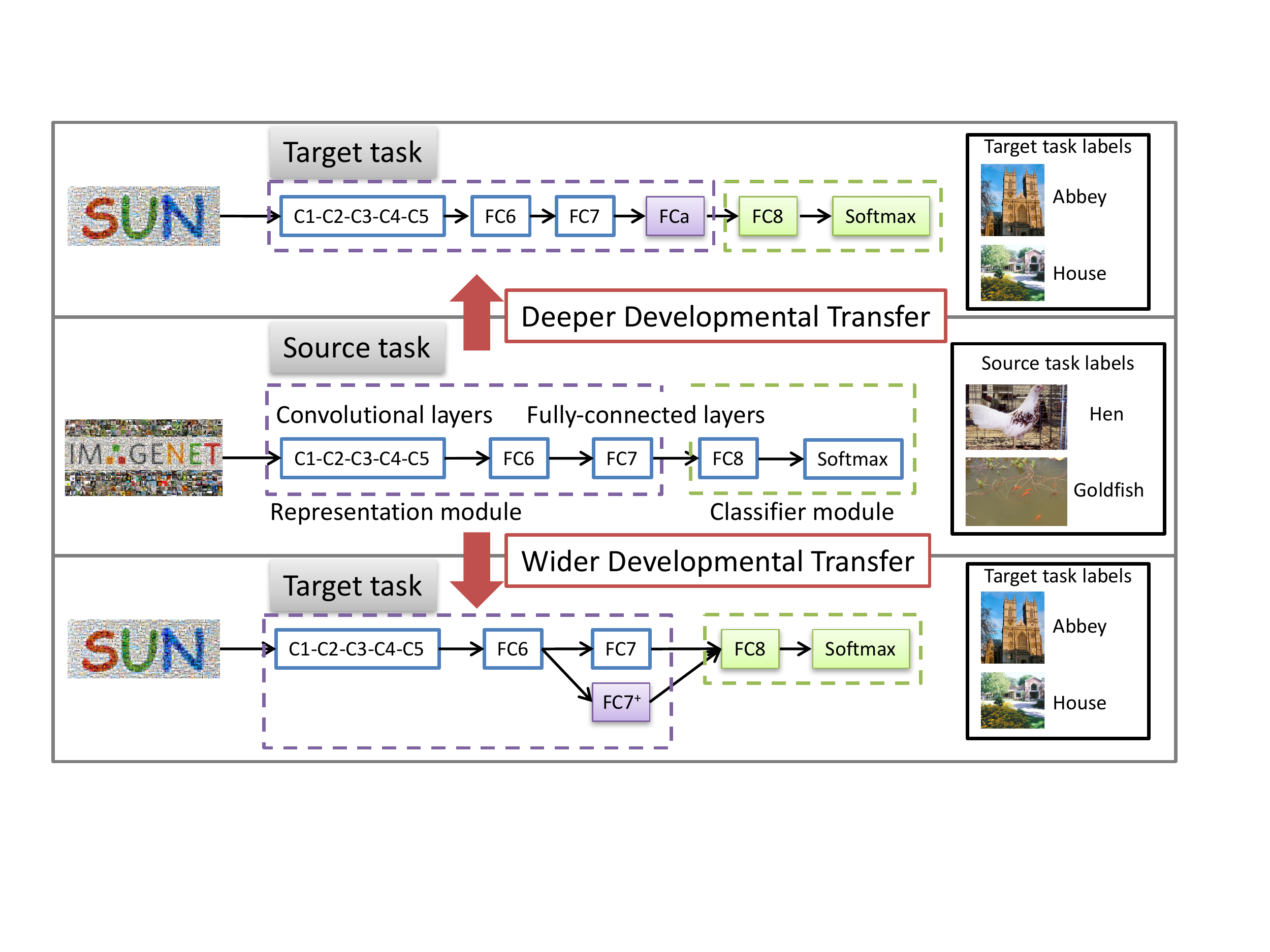}
\end{center}
\vspace{-0.5cm}
   \caption{Transfer and developmental learning of pre-trained CNNs by increasing model capacity for the recognition of novel categories from few examples. The network (\eg, AlexNet) is pre-trained on the source task (\eg, ImageNet classification) with abundant data (middle row). Different from the dominant paradigm of fine-tuning a fixed-capacity model, we {\em grow this network} when adapting it to a novel target task (\eg, SUN-$397$ scene classification) in two ways: (1) going deeper by adding more layers (top) and (2) going wider by adding more channels per layer (bottom).}
\label{fig:pipeline}
\vspace{-0.5cm}
\end{figure}
{\bf Developmental networks:} To address this issue, we explore ``developmental'' neural networks that grow in model capacity as new tasks as encountered. We demonstrate that growing a network, by adding additional units, facilitates knowledge transfer to new tasks. We explore two approaches to adding units as shown in Figure~\ref{fig:pipeline}: going deeper (more layers) and wider (more channels per layer). Through visualizations, we demonstrate that these additional units help guide the adaptation of pre-existing units. Deeper units allow for new compositions of pre-existing units, while wider units allow for the discovery of complementary cues that address the target task. Due to their progressive nature, developmental networks still remain accurate on their source task, implying that they can learn without forgetting. Finally, we demonstrate that developmental networks particularly facilitate continual transfer across multiple tasks. 

{\bf Developmental learning:}  Our approach is loosely inspired by developmental learning in cognitive science. Humans, and in particular children, have the remarkable ability to continually transfer previously-acquired knowledge to novel scenarios. Much of the literature from both neuroscience~\cite{nelson2001handbook} and psychology~\cite{huitt2003piaget} suggests that such sequential knowledge acquisition is intimately tied with a child's growth and development.

{\bf Contributions:}  Our contributions are three-fold. (1) We first demonstrate that the dominant paradigm of fine-tuning a fixed-capacity model is sub-optimal. (2) We explore several avenues for increasing model capacity, both in terms of going deeper (more layers) and wider (more channels per layer), and consistently find that increasing capacity helps, with a slight preference for widening. (3) We show that additional units must be normalized and scaled appropriately such that the ``pace of learning'' is balanced with existing units in the model. Finally, we use our analysis to build a relatively simple pipeline that ``grows'' a pre-trained model during fine-tuning, producing state-of-the-art results across a large number of standard and heavily benchmarked datasets (for scene classification, fine-grained recognition, and action recognition).

%-------------------------------------------------------------------------
\section{Related Work}
While there is a large body of work on transfer learning, much of it assumes a {\em fixed} capacity model 
~\cite{razavian2014cnn,azizpour2015factors,chubest2016best,zheng2016good,huh2016makes}.
Notable exceptions include~\cite{oquab2014learning}, who introduce an adaptation layer to facilitate transfer. Our work provides a systematic exploration of various methods for increasing capacity, including both the addition of new layers and widening of existing ones. Past work has explored strategies for preserving accuracy on the source task~\cite{li2016learning,furlanello2016active}, while our primary focus is on improving accuracy on the target task.
Most relevant to us are the {\em progressive} networks of~\cite{rusu2016progressive}, originally proposed for reinforcement learning. 
Interestingly,~\cite{rusu2016progressive,terekhov2015knowledge} focus on widening a target network to be twice as large as the source one, but fine-tune only the new units.
In contrast, we add a small fraction of new units (both by widening and deepening) but fine-tune {\em the entire network}, demonstrating that adaptation of old units is crucial for high performance. 

Transfer learning is related to both multi-task learning~\cite{razavian2014cnn,azizpour2015generic,oquab2014learning,girshick2014rich,gupta2016cross,tzeng2015simultaneous,misra2016cross,ando2005framework} and learning novel categories from few examples~\cite{wang2015model,koch2015siamese,lake2015human,santoro2016one,wang2016learning,li2016learning,bertinetto2016learning,hariharan2016low,wang2016ombining,vinyals2016matching,ravi2017optimization}. Past techniques have applied such approaches to transfer learning by learning networks that predict models rather than classes~\cite{wang2016learning,ravi2017optimization}. This is typically done without dynamically growing the number of parameters across new tasks (as we do).

In a broad sense, our approach is related to developmental learning~\cite{nelson2001handbook,huitt2003piaget,sigaud2016towards} and lifelong learning~\cite{thrun1998lifelong,mitchell2015never,tessler2016deep,pickett2016growing}. Different from the non-parametric shallow models (\eg, nearest neighbors) that increase capacity when memorizing new data~\cite{thrun1996learning,thrun1998clustering}, our developmental network cumulatively grows its capacity {\em from novel tasks}.

%-------------------------------------------------------------------------
\section{Approach Overview}
\label{sec:overview}
Let us consider a CNN architecture pre-trained on a source domain with abundant data, \eg, the vanilla AlexNet pre-trained on ImageNet (ILSVRC) with $1{,}000$ categories~\cite{krizhevsky2012imagenet,russakovsky2015imagenet}. We note  in Figure~\ref{fig:pipeline} that the CNN is composed of a feature representation module \begin{small}$\mathcal{F}$\end{small} (\eg, the five convolutional layers and two fully connected layers for AlexNet) and a classifier module \begin{small} $\mathcal{C}$\end{small} (\eg, the final fully-connected layer with \begin{small}$1{,}000$\end{small} units and the \begin{small}$1{,}000$\end{small}-way softmax for ImageNet classification). Transferring this CNN to a novel task with limited training data  (\eg, scene classification of  \begin{small}$397$\end{small} categories from SUN-$397$~\cite{xiao2016sun}) is typically done through fine-tuning~\cite{azizpour2015factors,agrawal2014analyzing,huh2016makes}.

In classic fine-tuning, the target CNN is instantiated and initialized as follows: (1) the representation module \begin{small}$\mathcal{F}_T$\end{small} is copied from \begin{small}$\mathcal{F}_S$\end{small} of the source CNN with the parameters \begin{small}$\Theta^\mathcal{F}_T=\Theta^\mathcal{F}_S$\end{small}; and (2) a new classifier model \begin{small}$\mathcal{C}_T$\end{small} (\eg, a new final fully-connected layer with \begin{small}$397$\end{small} units and the \begin{small}$397$\end{small}-way softmax for SUN-$397$ classification) is introduced with the parameters \begin{small}$\Theta^\mathcal{C}_T$\end{small} randomly initialized. All (or a portion of) the parameters \begin{small}$\Theta^\mathcal{F}_T$\end{small} and \begin{small}$\Theta^\mathcal{C}_T$\end{small} are fine-tuned by continuing the backpropagation, with a smaller learning rate for \begin{small}$\Theta^\mathcal{F}_T$\end{small}. Because \begin{small}$\mathcal{F}_T$\end{small} and\begin{small} $\mathcal{F}_S$\end{small} have identical network structure, the representational capacity is fixed during transfer. 

Our underlying thesis is that  fine-tuning will be facilitated by {\em increasing representational capacity} during transfer learning. We do so by adding \begin{small}$S$\end{small} new units \begin{small}$\{u_s\}_{s=1}^S$\end{small} into \begin{small}$\mathcal{F}_T$\end{small}.
As we will show later in our experiments, this significantly improves the ability to transfer knowledge to target tasks, particularly when fewer target examples are provided~\cite{tommasi2014learning}. We call our architecture a {\em developmental network}, in which the new representation module \begin{small}$\mathcal{F}_T^* = \mathcal{F}_T \cup \{u_s\}_{s=1}^S$\end{small}, and the classifier module remains \begin{small}$\mathcal{C}_T$\end{small}.

\begin{figure}[t]
\begin{center}
\centering
\begin{subfigure}[t]{0.24\textwidth}
        \centering
   \includegraphics[trim=.6cm 9cm 3.2cm 6cm,clip=true,width=.9\linewidth]{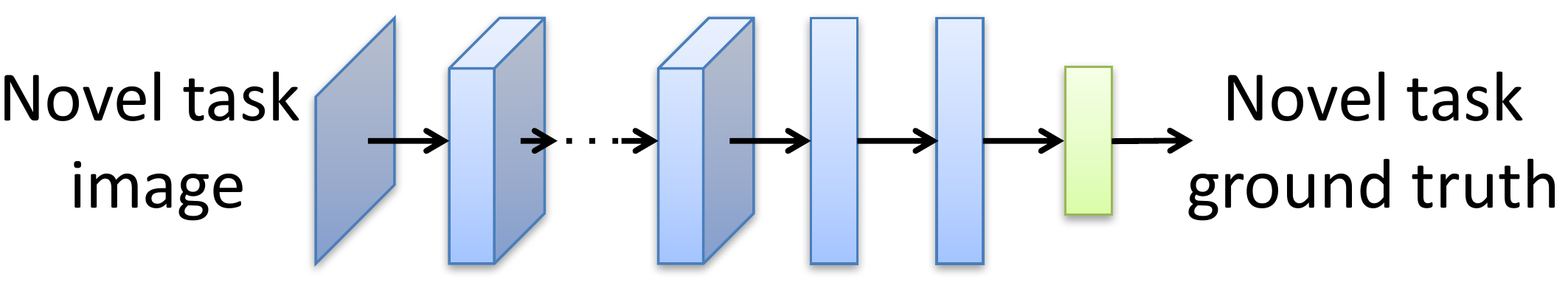}
        \caption{Classic Fine-Tuning}
    \end{subfigure}%
    ~  
\begin{subfigure}[t]{0.24\textwidth}
        \centering
   \includegraphics[trim=1.4cm 11cm 9cm 3cm,clip=true,width=.5\linewidth]{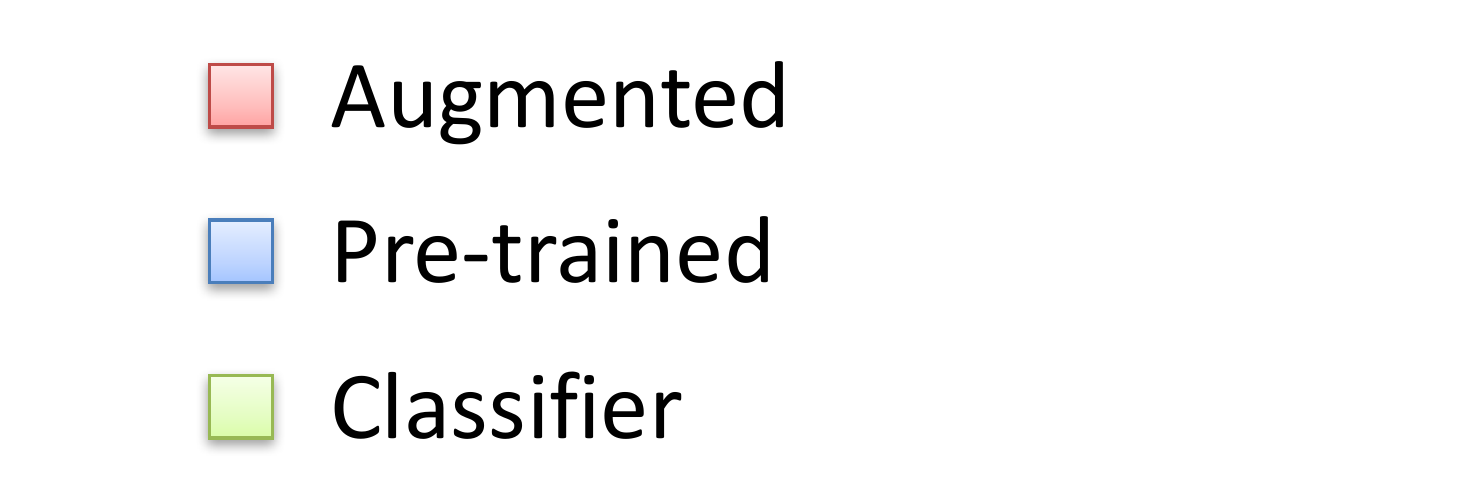}
    \end{subfigure}%
  
\begin{subfigure}[t]{0.24\textwidth}
        \centering
   \includegraphics[trim=.6cm 9cm 1.6cm 6cm,clip=true,width=.9\linewidth]{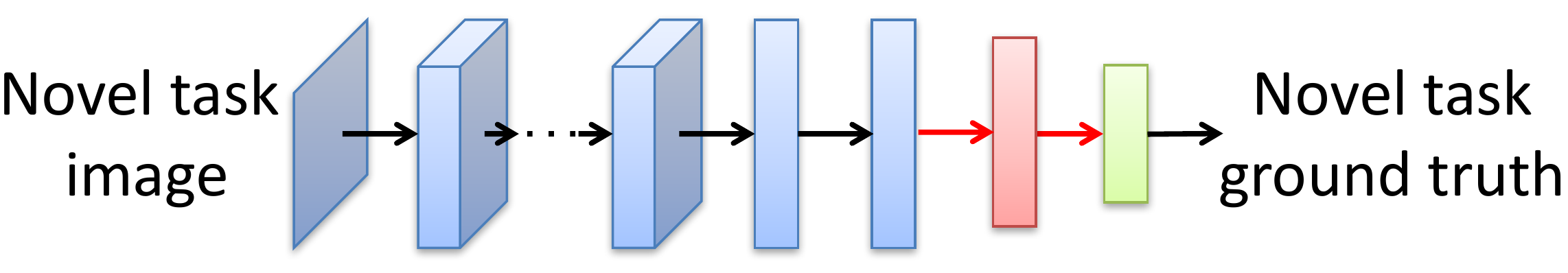}
        \caption{Depth Augmented Network (DA-CNN)}
        \label{fig:ddcnn}
    \end{subfigure}%
~    
    \begin{subfigure}[t]{0.24\textwidth}
        \centering
   \includegraphics[trim=.6cm 8.5cm 3.2cm 5cm,clip=true,width=.9\linewidth]{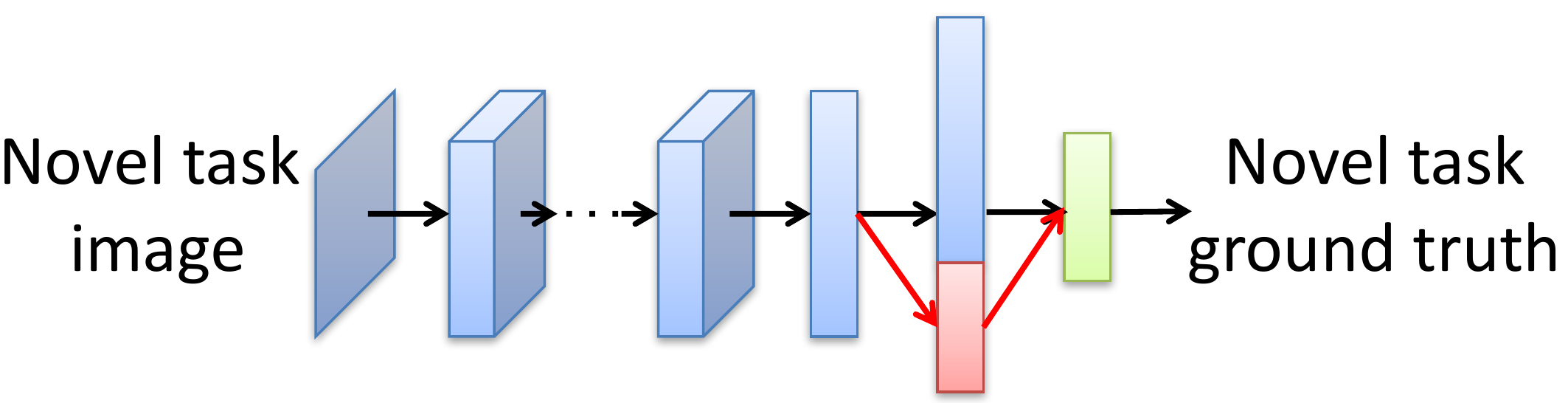}
        \caption{Width Augmented Network (WA-CNN)}
        \label{fig:wdcnn}
    \end{subfigure}%
   
\begin{subfigure}[t]{0.24\textwidth}
        \centering
   \includegraphics[trim=.6cm 8.5cm 1.5cm 5cm,clip=true,width=.9\linewidth]{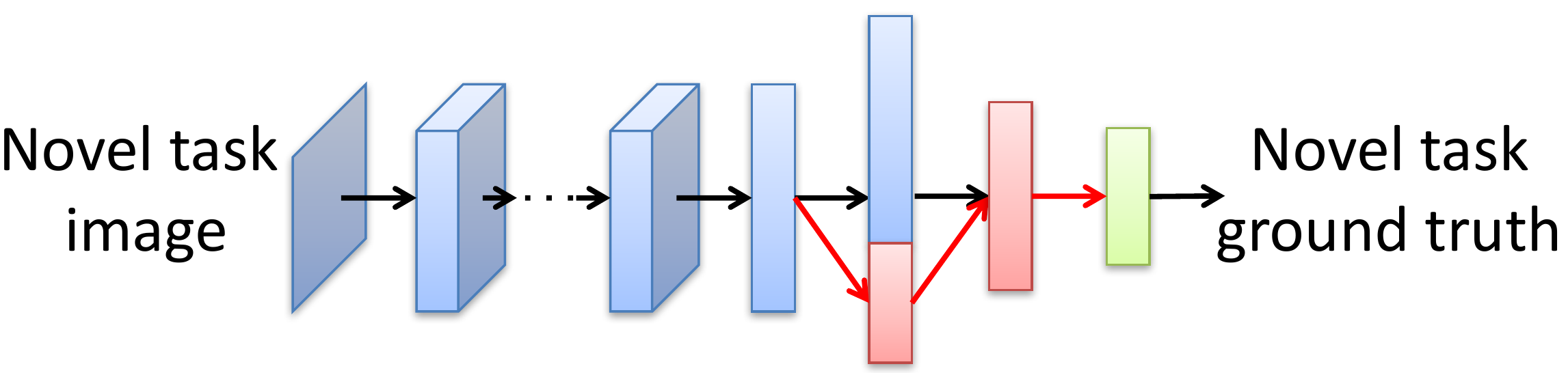}
        \caption{Jointly Depth and Width Augmented Network (DWA-CNN)}
        \label{fig:dwdcnn}
    \end{subfigure}%
~    
    \begin{subfigure}[t]{0.24\textwidth}
        \centering
   \includegraphics[trim=.6cm 8.5cm 3cm 5cm,clip=true,width=.9\linewidth]{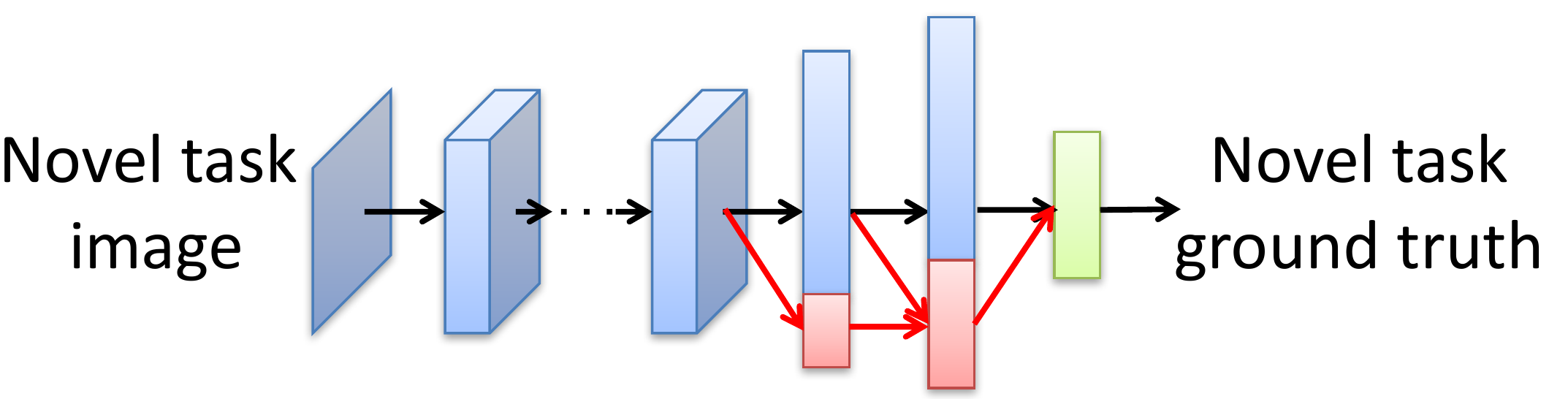}
        \caption{Recursively Width Augmented Network (WWA-CNN)}
        \label{fig:wwdcnn}
    \end{subfigure}

\end{center}
\vspace{-0.5cm}
   \caption{Illustration of classic fine-tuning (a) and variations of our developmental networks with augmented model capacity (b--e).}
\label{fig:network}
\vspace{-0.3cm}
\end{figure}

Conceptually, new units can be added to an existing network in a variety of ways. A recent analysis, however, suggests that early network layers tend to encode generic features, while later layers tend to endode task-specific features~\cite{yosinski2014transferable}. Inspired from this observation, we choose to explore new units at later layers. Specifically, we either construct a completely new top layer, leading to a {\em depth augmented network} (DA-CNN) as shown in Figure~\ref{fig:ddcnn}, or widen an existing top layer, leading to a {\em width augmented network} (WA-CNN) as shown in Figure~\ref{fig:wdcnn}.  We will explain these two types of network configurations in Section~\ref{sec:dnets}. Their combinations---a jointly depth and width augmented network (DWA-CNN) as shown in Figure~\ref{fig:dwdcnn} and a recursively width augmented network (WWA-CNN) as shown in Figure~\ref{fig:wwdcnn}---will also be discussed in Section~\ref{sec:exp}.

%-------------------------------------------------------------------------
\section{Developmental Networks}
\label{sec:dnets}
For the target task, let us assume that the representation module \begin{small}$\mathcal{F}_T$\end{small} with fixed capacity consists of \begin{small}$K$\end{small} layers \begin{small}$L_k, k =1, \ldots, K$\end{small} with hidden activations \begin{small}$\mathbfit{h}^k \in  \mathcal{R}^{n_k}$\end{small}, where \begin{small}${n_k}$\end{small} is the number of units at layer \begin{small}$k$\end{small}. Let  \begin{small}$\mathbfit{W}^k$\end{small} be the weights between layer \begin{small}$k$\end{small} and layer \begin{small}$k-1$\end{small}. That is, \begin{small}$\mathbfit{h}^k = f \left( {\mathbfit{W}^k\mathbfit{h}^{k-1}} \right)$\end{small}, where \begin{small}$f(\cdot)$\end{small} is a non-linear function, such as ReLU. For notational simplicity, \begin{small}$\mathbfit{h}^k$\end{small} already includes a constant \begin{small}$1$\end{small} as the last element and \begin{small}$\mathbfit{W}^k$\end{small} includes the bias terms.

\subsection{Depth Augmented Networks}
\label{sec:deeper}
A straightforward way to increase representational capacity is to construct a new top layer \begin{small}$L_a$\end{small} of size \begin{small}$S$\end{small} using \begin{small}$\{u_s\}_{s=1}^S$\end{small} on top of \begin{small}$L_K$\end{small}, leading to the depth augmented representation module \begin{small}$\mathcal{F}_T^*$\end{small} as shown in Figure~\ref{fig:ddcnn}. We view $L_a$ as an adaptation layer that 
allows for novel compositions of pre-existing units, thus avoiding dramatic modifications to the pre-trained layers  for their adaptation to the new task. The new activations \begin{small}$\mathbfit{h}^a = f \left( {\mathbfit{W}^a\mathbfit{h}^K} \right)$\end{small} in layer \begin{small}$L_a$\end{small} become the representation that is fed into the classifier module \begin{small}$\mathcal{C}_T$\end{small}, where \begin{small}$\mathbfit{W}^a$\end{small} denotes the weights between layers \begin{small}$L_a$\end{small} and \begin{small}$L_K$\end{small}.

%-------------------------------------------------------------------------
\subsection{Width Augmented Networks}
\label{sec:wider}
An alternative way is to expand the network by adding \begin{small}$\{u_s\}_{s=1}^S$\end{small} to some existing layers while keeping the depth of the network fixed as shown in Figure~\ref{fig:wdcnn}. Without loss of generality, we add all the units to the top layer \begin{small}$L_K$\end{small}. Now the new top representation layer \begin{small}$L_K^*$\end{small} consists of two blocks: the original \begin{small}$L_K$\end{small} and the added \begin{small}$L_K^+$\end{small} with units \begin{small}$\{u_s\}_{s=1}^S$\end{small}, leading to the width augmented representation module \begin{small}$\mathcal{F}_T^*$\end{small}. The connection weights between \begin{small}$L_K$\end{small} and the underneath layer \begin{small}$L_{K-1}$\end{small} remains, \ie, \begin{small}$\mathbfit{h}^K\!\! =\!\! f \left( {\mathbfit{W}^K\mathbfit{h}^{K-1}} \right)$\end{small}. We introduce additional lateral connection weights \begin{small}$\mathbfit{W}^{K^+}$\end{small} between \begin{small}$L_K^+$\end{small} and \begin{small}$L_{K-1}$\end{small}, which are randomly initialized, \ie, \begin{small}$\mathbfit{h}^{K^+}\!\! =\!\! f \left( {\mathbfit{W}^{K^+}\mathbfit{h}^{K-1}} \right)$\end{small}. Finally, the concatenated activations \begin{small}$\left[ { \mathbfit{h}^K,  \mathbfit{h}^{K^+}} \right]$\end{small} of size \begin{small}$n_K\!\!+\!\!S$\end{small} from layer \begin{small}$L_K^*$\end{small} are fed into the classifier module.

\subsection{Learning at the Same Pace} 
\label{sec:scaling}
Ideally, our hope is that the new and old units cooperate with each other to boost the target performance. For width augmented networks, however, the units start to learn at a different pace during fine-tuning: while the original units at layer \begin{small}$L_k$\end{small} are already well learned on the source domain and only need a small modification for adaptation, the new set of units at layer \begin{small}$L_k^+$\end{small} are just set up through random initialization. They thus have disparate learning behaviors, in the sense that their activations generally have different scales. Na\"{i}vely concatenating these activations would restrict the corresponding units, leading to degraded performance and even causing collapsed networks, since the larger activations dominate the smaller ones~\cite{liu2016parsenet}. Although the weights might adjust accordingly as fine-tuning processes, they require very careful initialization and tuning of parameters, which is dataset dependent and thus not robust. This is partially the reason that the previous work showed that network expansion was inferior to standard fine-tuning~\cite{li2016learning}.

To reconcile the learning pace of the new and pre-existing units, we introduce an additional normalization and adaptive scaling scheme in width augmented networks, which is inspired by the recent work on combining multi-scale pre-trained CNN features from different layers~\cite{liu2016parsenet}. More precisely, after weight initialization of \begin{small}$\mathcal{F}_T^*$\end{small}, we first apply an \begin{small}$\mathcal{L}_2$\end{small}-norm normalization to the activations $\mathbfit{h}^k$ and $\mathbfit{h}^{k^+}$, respectively:
\begin{small}
\begin{align}
\label{eq:norm}
{\widehat {\mathbfit{h}}^k} =\left. {{{\mathbfit{h}^k}}}\middle/{{\left\| {{\mathbfit{h}^k}} \right\|}}_2\right. &,\enskip {\widehat {\mathbfit{h}}^{k^+}} =\left. {{{\mathbfit{h}^{k^+}}}}\middle/ {{\left\| {{\mathbfit{h}^{k^+}}} \right\|}}_2\right. .
\end{align}
\end{small}By normalizing these activations, their scales become homogeneous. Simply normalizing the norms to \begin{small}$1$\end{small} slows down the learning and makes it hard to train the network, since the features become very small. Consistent with~\cite{liu2016parsenet}, we normalize them to a larger value (\eg, \begin{small}$10$\end{small} or \begin{small}$20$\end{small}), which encourages the network to learn well. We then introduce a scaling parameter \begin{small}$\gamma$\end{small} for each channel to scale the normalized value:
\begin{small}
\begin{align}
\label{eq:scale}
y_i^k = {\gamma _i}\widehat h_i^k,\enskip y_j^{k^+} = {\gamma _j}\widehat h_j^{k^+}.
\end{align}
\end{small}We found that for depth augmented networks, while this additional stage of normalization and scaling is not crucial, it is still beneficial. In addition, this stage only introduces negligible extra parameters, whose number is equal to the total number of channels. During fine-tuning, the scaling factor $\gamma$ is fine-tuned by backpropagation as in~\cite{liu2016parsenet}.

%-------------------------------------------------------------------------
\section{Experimental Evaluation}
\label{sec:exp}
In this section, we explore the use of our developmental networks for transferring a pre-trained CNN to a number of supervised learning tasks with insufficient data, including scene classification, fine-grained recognition, and action recognition. We begin with extensive evaluation of our approach on scene classification of the SUN-$397$ dataset, focusing on the variations of our networks and different design choices. We also show that the network remains accurate on the source task. We then provide an in-depth analysis of fine-tuning procedures to qualitatively understand why fine-tuning with augmented network capacity outperforms classic fine-tuning. We further evaluate our approach on other novel categories and compare with state-of-the-art approaches. Finally, we investigate whether progressive augmenting outperforms fine-tuning a fixed large network and investigate how to cumulatively add new capacity into the network when it is gradually adapted to multiple tasks.

\textbf{Implementation details:} Following the standard practice, for computational efficiency and easy fine-tuning we use the Caffe~\cite{jia2014caffe} implementation of AlexNet~\cite{krizhevsky2012imagenet}, pre-trained on ILSVRC 2012~\cite{russakovsky2015imagenet}, as our reference network in most of our experiments. We found that our observations also held for other network architectures. We also provide a set of experiment using VGG16~\cite{simonyan2015very}. For the target tasks, we randomly initialize the classifier layers and our augmented layers. During fine-tuning, after resizing the image to be \begin{small}$256 \times 256$\end{small}, we generate the standard augmented data including random crops and their flips as implemented in Caffe~\cite{jia2014caffe}. During testing, we only use the central crop, unless otherwise specified. For a fair comparison, fine-tuning is performed using stochastic gradient descent (SGD) with the ``step'' learning rate policy, which drops the learning rate in steps by a factor of  \begin{small}$10$\end{small}.  The new layers are fine-tuned at a learning rate \begin{small}$10$\end{small} times larger than that of the pre-trained layers (if they are fine-tuned). We use standard momentum  \begin{small}$0.9$\end{small} and weight decay \begin{small}$0.0005$\end{small} without further tuning.

\subsection{Evaluation and Analysis on SUN-397}
\label{sec:sun397}
We start our evaluation on scene classification of the SUN-$397$ dataset, a medium-scale dataset with around \begin{small}$108K$\end{small} images and \begin{small}$397$\end{small} classes~\cite{xiao2016sun}. In contrast to other fairly small-scale target datasets, SUN-$397$ provides sufficient number of categories and examples while demonstrating apparent dissimilarity with the source ImageNet dataset. This greatly benefits our insight into fine-tuning procedures and leads to clean comparisons under controlled settings.

We follow the experimental setup in~\cite{agrawal2014analyzing,huh2016makes}, which uses a nonstandard train/test split since it is computationally expensive to run all of our experiments on the \begin{small}$10$\end{small} standard subsets proposed by~\cite{xiao2016sun}. Specifically, we randomly split the dataset into train, validation, and test parts using \begin{small}$50\%$\end{small}, \begin{small}$10\%$\end{small}, and \begin{small}$40\%$\end{small} of the data, respectively. The distribution of categories is uniform across all the three sets. We report \begin{small}$397$\end{small}-way multi-class classification accuracy averaged over all categories, which is the standard metric for this dataset. We report the results using a single run due to computational constraints. Consistent with the results reported in~\cite{agrawal2014analyzing,huh2016makes}, the standard deviations of accuracy on SUN-$397$ classification are negligible, and thus having a single run should not affect the conclusions that we draw. For a fair comparison, fine-tuning is performed for around \begin{small}$60$\end{small} epochs using SGD with an initial learning rate of \begin{small}$0.001$\end{small}, which is reduced by a factor of \begin{small}$10$\end{small} around every \begin{small}$25$\end{small} epochs. All the other parameters are the same for all approaches. 
\begin{table}
\small
\begin{center}
\resizebox{\linewidth}{!}{
\begin{tabular}{c|c| c|c|c|c|c}
\hline
\multirow{2}{*}{Network}&\multirow{2}{*}{Type}& \multirow{2}{*}{Method} & \multicolumn{4}{c}{Acc (\%)} \\ \cline{4-7}
& & & New & $FC_7$--New & $FC_6$--New & All \\ \hline
\multirow{6}{*}{AlexNet}& \multirow{2}{*}{\tabincell{c}{Baselines}} & Finetuning-CNN & 53.63 & 54.75 & 54.29 & 55.93\\ 
& &~\cite{agrawal2014analyzing,huh2016makes} & 48.4 & --- & 51.6 & 52.2\\ \cline{2-7}
&\multirow{2}{*}{\tabincell{c}{Single \\(Ours)}} & DA-CNN & 54.24 & 56.48 & 57.42 & 58.54 \\ 
 & &WA-CNN & \bf{56.81} & 56.99 & 57.84 & 58.95 \\ \cline{2-7}
& \multirow{2}{*}{\tabincell{c}{Combined \\(Ours)}} & DWA-CNN & 56.07 & 56.41 & 56.97 & 57.75 \\
 & &WWA-CNN & 56.65 &\bf{57.10} & \bf{58.16} & \bf{59.05} \\ \hline
\hline
\multirow{3}{*}{VGG16} &{Baselines} & Finetuning-CNN &60.77 & 59.09 & 50.54 & 62.80\\ \cline{2-7}
&\multirow{2}{*}{\tabincell{c}{Single \\(Ours)}} & DA-CNN & 61.21 & 62.85& 63.07 & 65.55 \\
& &WA-CNN & \bf{63.61} & \bf{64.00} & \bf{64.15} & \bf{66.54} \\ \hline
\end{tabular}
}
\vspace{-0.2cm}
\caption{Performance comparisons of classification accuracy (\%) between the variations of our developmental networks {\em with augmented model capacity} and classic fine-tuning {\em with fixed model capacity} on scene classification of the SUN-$397$ dataset. The variations include: (1) for AlexNet, depth augmented network (DA-CNN), width augmented network (WA-CNN), jointly depth and width augmented network (DWA-CNN), and recursively width augmented network (WWA-CNN); and (2) for VGG16, DA-CNN and WA-CNN. Both our networks and the baselines are evaluated in four scenarios of gradually increasing the degree of fine-tuning, including fine-tuning only new layers, from $FC_7$ to new layers, from $FC_6$ to new layers, and the entire network. Ours significantly outperform the vanilla fine-tuned CNN in all these scenarios.}
\label{tab:sunmain}
\end{center}
\vspace{-0.7cm}
\end{table}

\textbf{Learning with augmented network capacity:} We first evaluate our developmental networks obtained by introducing a {\em single} new layer to deepen or expand the pre-trained AlexNet. For the depth augmented network (DA-CNN), we add a new fully connected layer \begin{small}$FCa$\end{small} of size \begin{small}$S^{D}$\end{small} on top of \begin{small}$FC_7$\end{small} whose size is \begin{small}$4{,}096$\end{small}, where \begin{small}$S^{D} \in \{ 1{,}024, 2{,}048, 4{,}096, 6{,}144 \}$\end{small}. For the width augmented network (WA-CNN), we add a set of \begin{small}$S^{W}$\end{small} new units as \begin{small}$FC^+_7$\end{small} to \begin{small}$FC_7$\end{small}, where \begin{small}$S^{W} \in \{ 1{,}024, 2{,}048 \}$\end{small}. After their structures are adapted to the target task, the networks then continue learning in four scenarios of gradually increasing the degree of fine-tuning: (1) ``New'': we only fine-tune the new layers, including the classifier layers and the augmented layers, while freezing the other pre-trained layers (\ie, the off-the-shelf use case of CNNs); (2) ``\begin{small}$FC_7$\end{small}--New'': we fine-tune from the \begin{small}$FC_7$\end{small} layer; (3) ``\begin{small}$FC_6$\end{small}--New'': we fine-tune from the \begin{small}$FC_6$\end{small} layer; (4) ``All'': we fine-tune the entire network.

Table~\ref{tab:sunmain} summarizes the performance comparison with classic fine-tuning. The performance gap between our implementation of the fine-tuning baseline and that in~\cite{agrawal2014analyzing,huh2016makes} is mainly due to different number of iterations: we used twice of the number of epochs in ~\cite{agrawal2014analyzing,huh2016makes} (\begin{small}$30$\end{small} epochs), leading to improved accuracy. Note that these numbers cannot be directly compared against other publicly reported results due to different data split. With relatively sufficient data, fine-tuning through the full network yields the best performance for all the approaches. Both our DA-CNN and WA-CNN significantly outperform the vanilla fine-tuned CNN {\em in all the different fine-tuning scenarios}. This verifies the effectiveness of increasing model capacity when adapting it to a novel task. While they have achieved comparable performance, WA-CNN slightly outperforms DA-CNN.

\begin{table}[t]
\small
\begin{center}
\resizebox{\linewidth}{!}{
\begin{tabular}{c| c|c|c|c|c}
\hline
Method & Configuration & New & $FC_7$--New & $FC_6$-New & All \\ \hline
\multirow{4}{*}{\tabincell{c}{DA-\\CNN}} & $FC_a$--$1{,}024$ & 53.36 & 56.31 & 57.22 & 57.98 \\
&$FC_a$--$2{,}048$ & 53.82 & 56.47 & 57.14 & 58.07 \\
&$FC_a$--$4{,}096$ & 54.02 & 56.46 & 57.41 & 58.32 \\
&$FC_a$--$6{,}144$ & 54.24 & 56.48 & 57.42 & 58.54 \\ \hline
\hline
\multirow{2}{*}{\tabincell{c}{WA-\\CNN}} & $FC_7^{+}$--$1{,}024$ & 56.46 & 56.71 & 57.55 & 58.90 \\
&$FC_7^+$--$2{,}048$ & 56.81 & 56.99 & 57.84 & 58.95 \\ \hline
\hline
\multirow{2}{*}{\tabincell{c}{DWA-\\CNN}} & $FC_7^+$--$1{,}024$--$FC_a$--$1{,}024$ & 55.44 & 55.77 & 56.71 & 57.49 \\
&$FC_7^+$--$2{,}048$-$FC_a$--$2{,}048$ & 56.07 & 56.41 & 56.97 & 57.75 \\ \hline
\hline
\multirow{3}{*}{\tabincell{c}{WWA-\\CNN}} & $FC_6^+$--$512$--$FC_7^+$--$1{,}024$ & 56.13 & 57.10 & 57.65 & 58.80 \\ 
& $FC_6^+$--$1{,}024$--$FC_7^+$--$2{,}048$ & 56.49 & 57.10 & 57.98 &  59.05 \\
& $FC_6^+$--$2{,}048$--$FC_7^+$--$4{,}096$ & 56.65 & 57.03 & 58.16 & 58.98 \\ \hline
\end{tabular}
}
\vspace{-0.2cm}
\caption{Diagnostic analysis of classification accuracy (\%) for the variations of our developmental networks with different number of new units on SUN-$397$.}
\label{tab:detail}
\end{center}
\vspace{-0.7cm}
\end{table}

\textbf{Increasing network capacity through combination or recursion:} Given the promise of DA-CNN and WA-CNN, we further augment the network by making it both deeper and wider or two-layer wider. For the jointly depth and width augmented network (DWA-CNN) (Figure~\ref{fig:dwdcnn}), we add \begin{small}$FC_a$\end{small} of size \begin{small}$S^{DW}$\end{small} on top of \begin{small}$FC_7$\end{small} while expanding \begin{small}$FC_7$\end{small} using \begin{small}$FC^+_7$\end{small} of size \begin{small}$S^{DW}$\end{small}, where \begin{small}$S^{DW} \in\!\! \{ 1{,}024, 2{,}048\}$\end{small}. For the recursively width augmented network (WWA-CNN) (Figure~\ref{fig:wwdcnn}), we both expand \begin{small}$FC_7$\end{small} using \begin{small}$FC^+_7$\end{small} of size \begin{small}$S^{WW}_7$\end{small} and \begin{small}$FC_6$\end{small} using \begin{small}$FC^+_6$\end{small} of size \begin{small}$S^{WW}_6$\end{small}, where \begin{small}$S^{WW}_7\!\! \in \!\!\{ 1{,}024, 2{,}048, 4{,}096\}$\end{small} and  \begin{small}$S^{WW}_6$\end{small} is half of \begin{small}$S^{WW}_7$\end{small}. 

We compare DWA-CNN and WWA-CNN with DA-CNN and WA-CNN in Table~\ref{tab:sunmain}. The two-layer WWA-CNN generally achieves the best performance, indicating the importance of augmenting model capacity at different and complementary levels. The jointly DWA-CNN lags a little bit behind the purely WA-CNN. This implies different learning behaviors when we make the network deeper or wider. Their combination thus becomes a non-trivial task.

\textbf{Diagnostic analysis:} While we summarize the best performance in Table~\ref{tab:sunmain}, a diagnostic experiment in Table~\ref{tab:detail} on the number of augmented units \begin{small}$S^{D}$\end{small}, \begin{small}$S^{W}$\end{small}, \begin{small}$S^{DW}$\end{small}, and \begin{small}$S^{WW}$\end{small} shows that {\em all of these variations of network architectures significantly outperform classic fine-tuning}, indicating the robustness of our approach. We found that this observation was also consistent with other datasets, which we evaluated in the later section. Overall, the performance increases with the augmented model capacity (represented by the size of augmented layers), although the performance gain diminishes with the increasing number of new units.

\textbf{Importance of reconciling the learning pace of new and old units:} The previous work showed that network expansion did not introduce additional benefits~\cite{li2016learning}. We argue that its unsatisfactory performance is because of the failure of taking into account the different learning pace of new and old units. After exploration of different strategies, such as initialization, we found that the performance of a width augmented network significantly improves by a simple normalization and scaling scheme when concatenating the pre-trained and expanded layers. This issue is investigated for both types of model augmentation in Table~\ref{tab:scaling}. The number of new units is generally \begin{small}$2{,}048$\end{small}; in the case of copying weights of the pre-trained \begin{small}$FC_7$\end{small} and then adding random noises as initialization for \begin{small}$FC_7^+$\end{small}, we use \begin{small}$4{,}096$\end{small} new units.
\begin{table}[t]
\small
\begin{center}
\resizebox{\linewidth}{!}{
\begin{tabular}{c| c|c|c|c|c}
\hline
Method & Scaling & New & $FC_7$--New & $FC_6$--New & All \\ \hline
\multirow{2}{*}{\tabincell{c}{DA-CNN\\$FC_a$--$2{,}048$}} & \textit{w/o} & \bf{53.82} & \bf{56.47} & 56.25 & 57.21 \\
&\textit{w/} & 53.51 & 56.15 & \bf{57.14} & \bf{58.07}\\ \hline
\hline
\multirow{3}{*}{\tabincell{c}{WA-CNN\\$FC_7^+$--$2{,}048$)}} & \textit{w/o} (rand)& 53.78  & 54.66 & 49.72 & 51.34 \\ 
&\textit{w/o} (copy+rand)& 53.62 & 54.35 & 53.70 & 55.31 \\
&\textit{w/} & \bf{56.81} & \bf{56.99} & \bf{57.84} & \bf{58.95}	\\ \hline
\end{tabular}
}
\vspace{-0.2cm}
\caption{Performance comparisons of classification accuracy (\%) for our depth (DA-CNN) or width (WA-CNN) augmented network {\em with and without} introducing normalization and scaling on SUN-$397$. The number of new units in $FC_a$ for DA-CNN or in $FC^+_7$ for WA-CNN is generally $2{,}048$. Our normalization and scaling strategy reconciles the learning pace of new and old units, and thus greatly benefits both types of networks, {\em in particular WA-CNN}.}
\label{tab:scaling}
\end{center}
\vspace{-0.7cm}
\end{table}

For WA-CNN, if we na\"{\i}vely add new units without considering scaling, Table~\ref{tab:scaling} shows that the performance is either only marginally better or even worse than classic fine-tuning (when fine-tuning more aggressively) in Table~\ref{tab:sunmain}. This is consistent with the observation made in~\cite{li2016learning}. However, once the learning pace of the new and old units is re-balanced by scaling, WA-CNN exceeds the baseline by a large margin. For DA-CNN, directly adding new units without scaling already greatly outperforms the baseline, which is consistent with the observation in~\cite{oquab2014learning}, although scaling provides additional performance gain. This suggests slightly different learning behaviors for depth and width augmented networks. When a set of new units are added to form a purely new layer, they have relatively more freedom to learn from scratch, making the additional scaling {\em beneficial yet inessential}. When the units are added to expand a pre-trained layer, however, the constraints from the synergy require them to learn to collaborate with the pre-existing units, which is explicitly achieved by the additional scaling.

\textbf{Evaluation with the VGG16 architecture:} Table~\ref{tab:sunmain} also summarizes the performance of DA-CNN and WA-CNN using VGG16~\cite{simonyan2015very} and shows the generality of our approach. Due to GPU memory and time constraints, we reduce the batch size and perform fine-tuning for around \begin{small}$30$\end{small} epochs using SGD. All the other parameters are the same as before. Also, following the standard practice in Fast R-CNN~\cite{girshick2015fast}, we fine-tune from the layer ${\mathcal Conv}2\_1$ in the ``All'' scenario.

\textbf{Learning without forgetting:} Conceptually, due to their developmental nature, our networks should remain accurate on their source task. Table~\ref{tab:source} validates such ability of {\em learning without forgetting} by showing their classification performance on the source ImageNet dataset.
\begin{table}[h]
\begin{center}
\small
\begin{tabular}{c| c | c}
  \clineB{1-3}{2.5}
Type & Method & Acc (\%) \\ \hline
\multirow{1}{*}{{\em Oracle}} & {\em ImageNet-AlexNet} & {\em 56.9}\\ \hline \hline
\multirow{2}{*}{References} 
&LwF~\cite{li2016learning} & 55.9 \\
&Joint~\cite{li2016learning} & 56.4 \\
\hline \hline
\multirow{2}{*}{\tabincell{c}{Ours}} & DA-CNN & 55.3 \\
& WA-CNN & 51.5\\ 
  \clineB{1-3}{2.5}
\end{tabular}
\caption{Demonstration of the ability of {\em learning without forgetting} on the source (ImageNet) ILSVRC $2012$ validation set. For our DA-CNN and WA-CNN that are fine-tuned on SUN-$397$, we re-fine-tune on the source ILSVRC $2012$ training set, \ie, re-training a new $1{,}000$-way classifier layer and fine-tuning the augmented layers. We show the results of the oracle (\ie, the original AlexNet) and the approaches that are {\em specifically designed} to preserve the performance on the source task during transfer~\cite{li2016learning} as references. While our approach focuses on improving the performance on the target task, it remains accurate on the source task. In addition, the existing approaches~\cite{li2016learning} can be naturally incorporated into our approach to further improve the performance on both source and target tasks.}
\label{tab:source}
\end{center}
\end{table}

\vspace{-.9cm}
\subsection{Understanding of Fine-Tuning Procedures}
\label{sec:visual}
We now analyze the fine-tuning procedures from various perspectives to gain insight into how fine-tuning modifies the pre-trained network and why it helps by increasing model capacity. We evaluate on the SUN-$397$ validation set. For a clear analysis and comparison, we focus on DA-CNN and WA-CNN, both with \begin{small}${2{,}048}$\end{small} new units.
\begin{figure}[t]
\begin{center}
\centering
\begin{subfigure}[t]{0.24\textwidth}
        \centering
   \includegraphics[trim=2.2cm 2.5cm 2cm 1.5cm,clip=true,width=.82\linewidth]{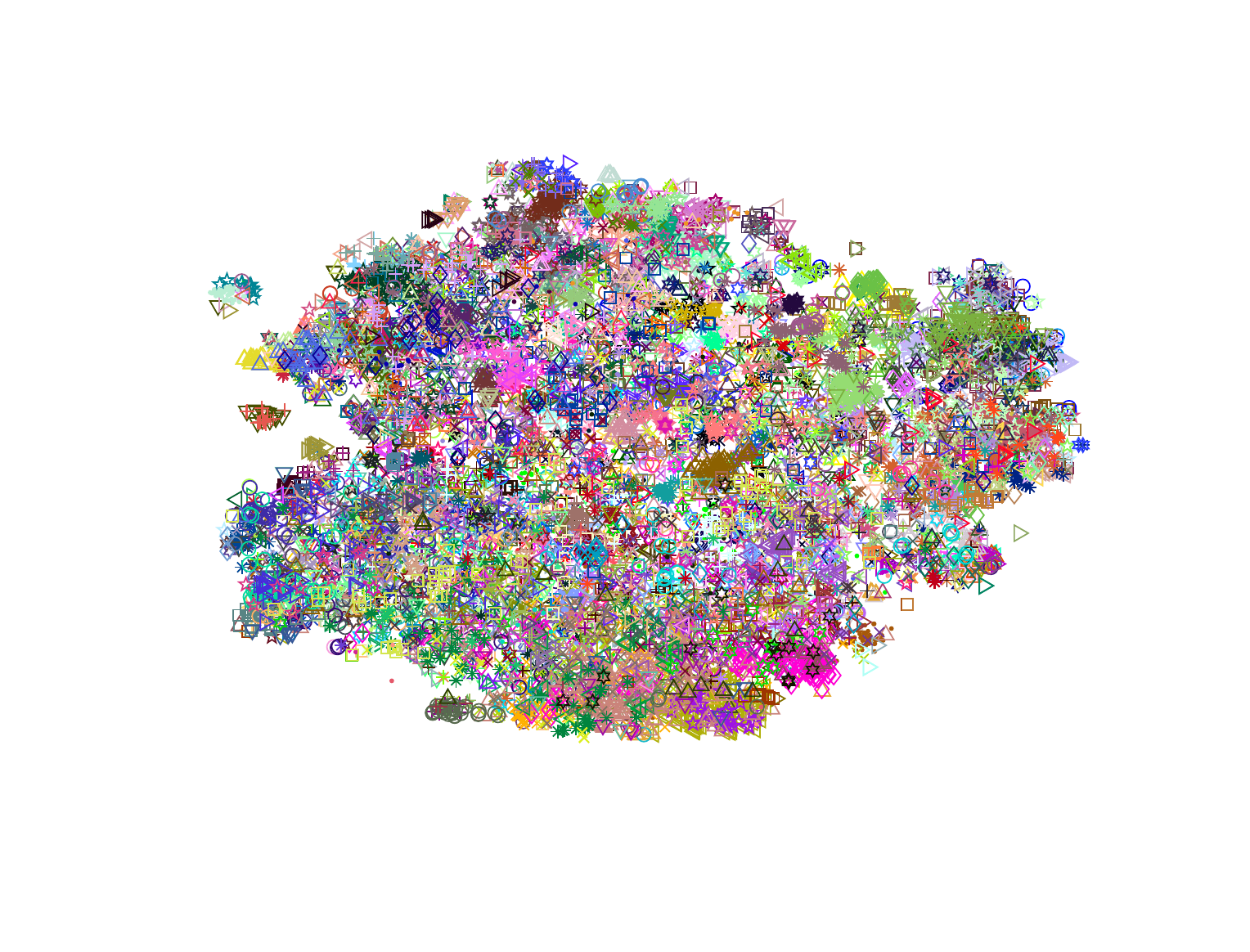}
        \caption{Pre-Trained Network}
    \end{subfigure}%
    ~  
\begin{subfigure}[t]{0.24\textwidth}
        \centering
   \includegraphics[trim=2.2cm 1cm 2cm 2cm,clip=true,width=.82\linewidth]{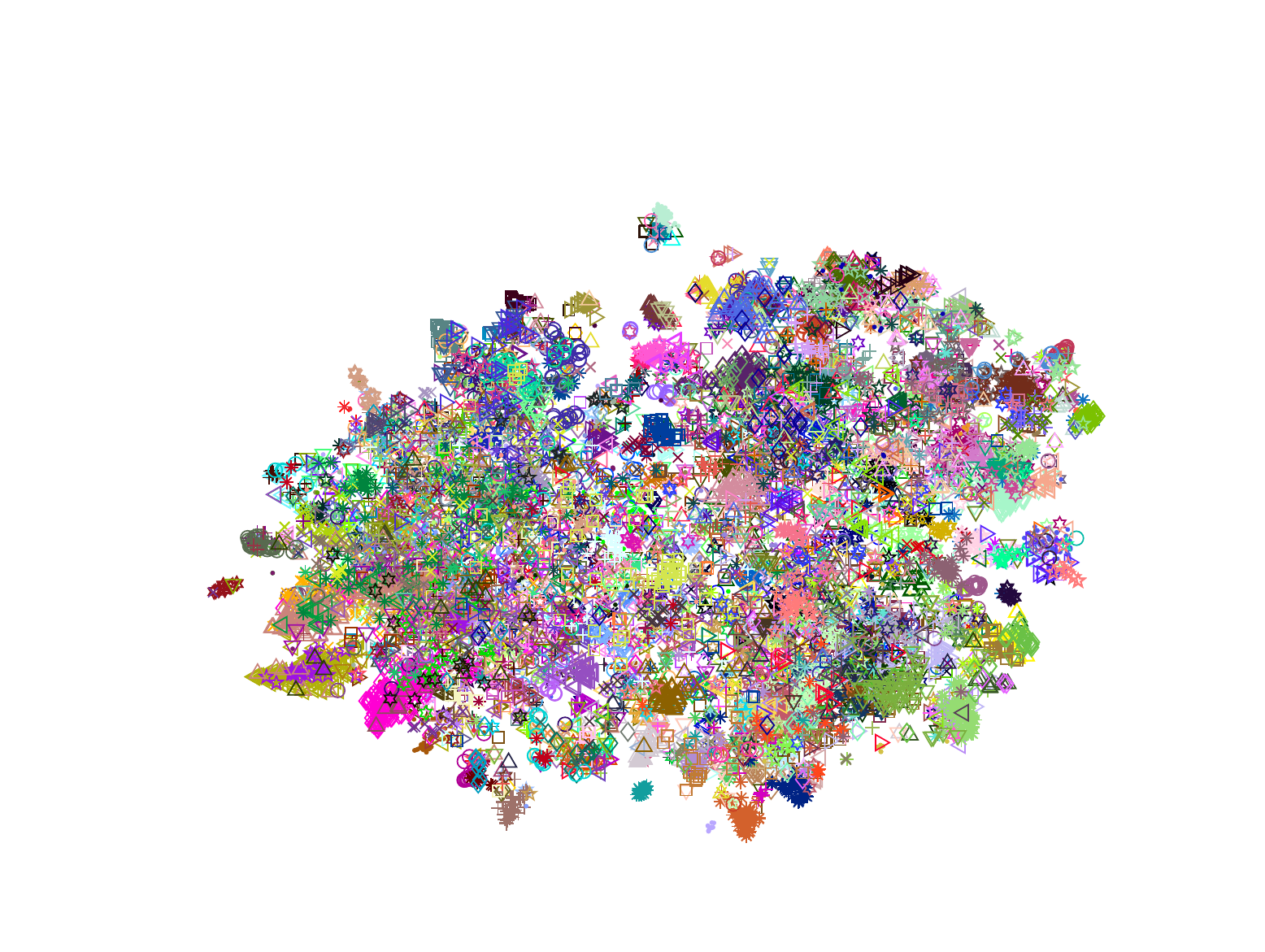}
        \caption{Classic Fine-Tuning}
    \end{subfigure}%
    
    \begin{subfigure}[t]{0.24\textwidth}
        \centering
   \includegraphics[trim=2.2cm 2cm 2cm 2cm,clip=true,width=.82\linewidth]{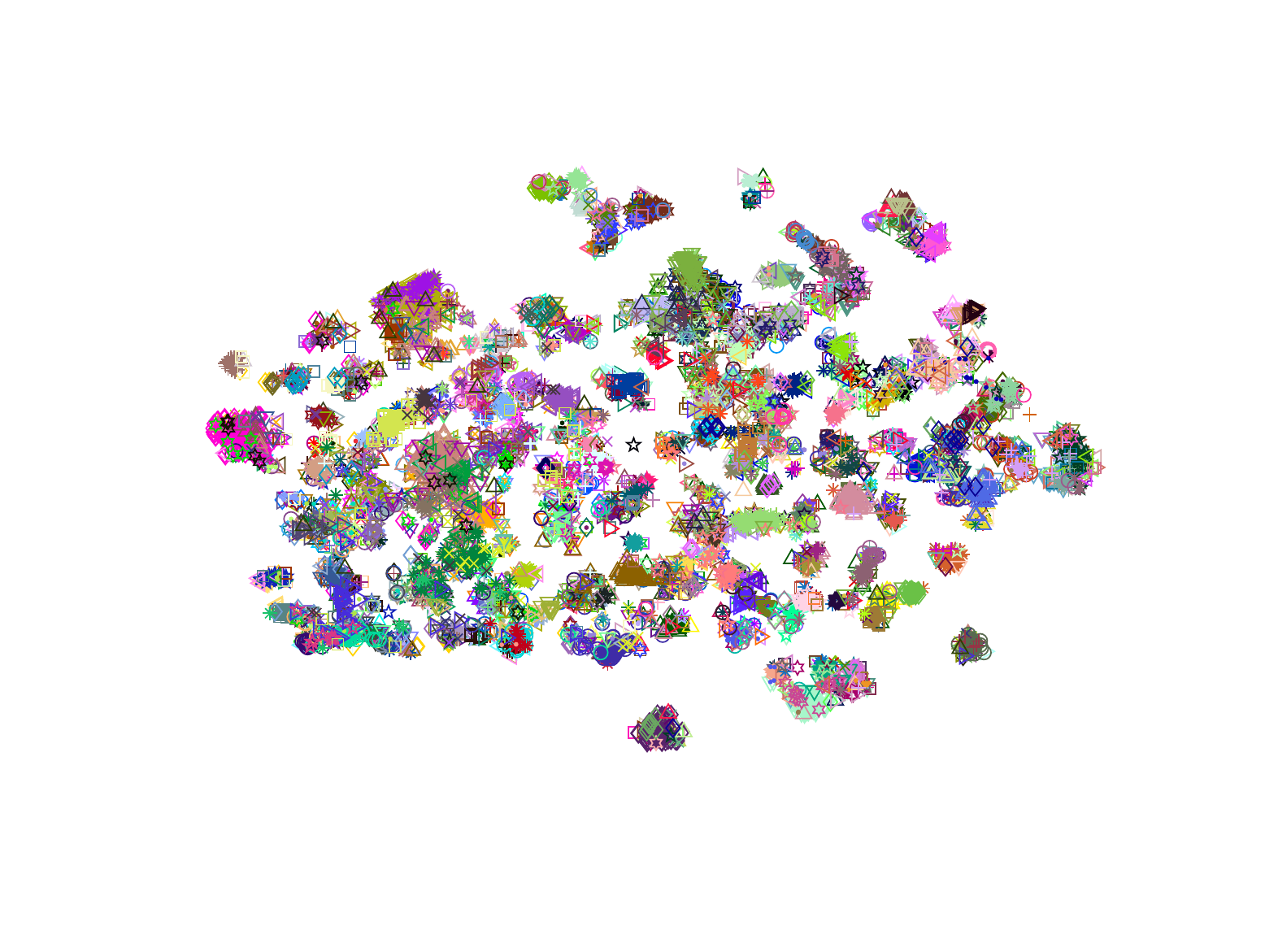}
        \caption{Depth Augmented Network (DA-CNN)}
    \end{subfigure}%
    ~  
\begin{subfigure}[t]{0.24\textwidth}
        \centering
   \includegraphics[trim=2.2cm 2cm 2cm 2cm,clip=true,width=.82\linewidth]{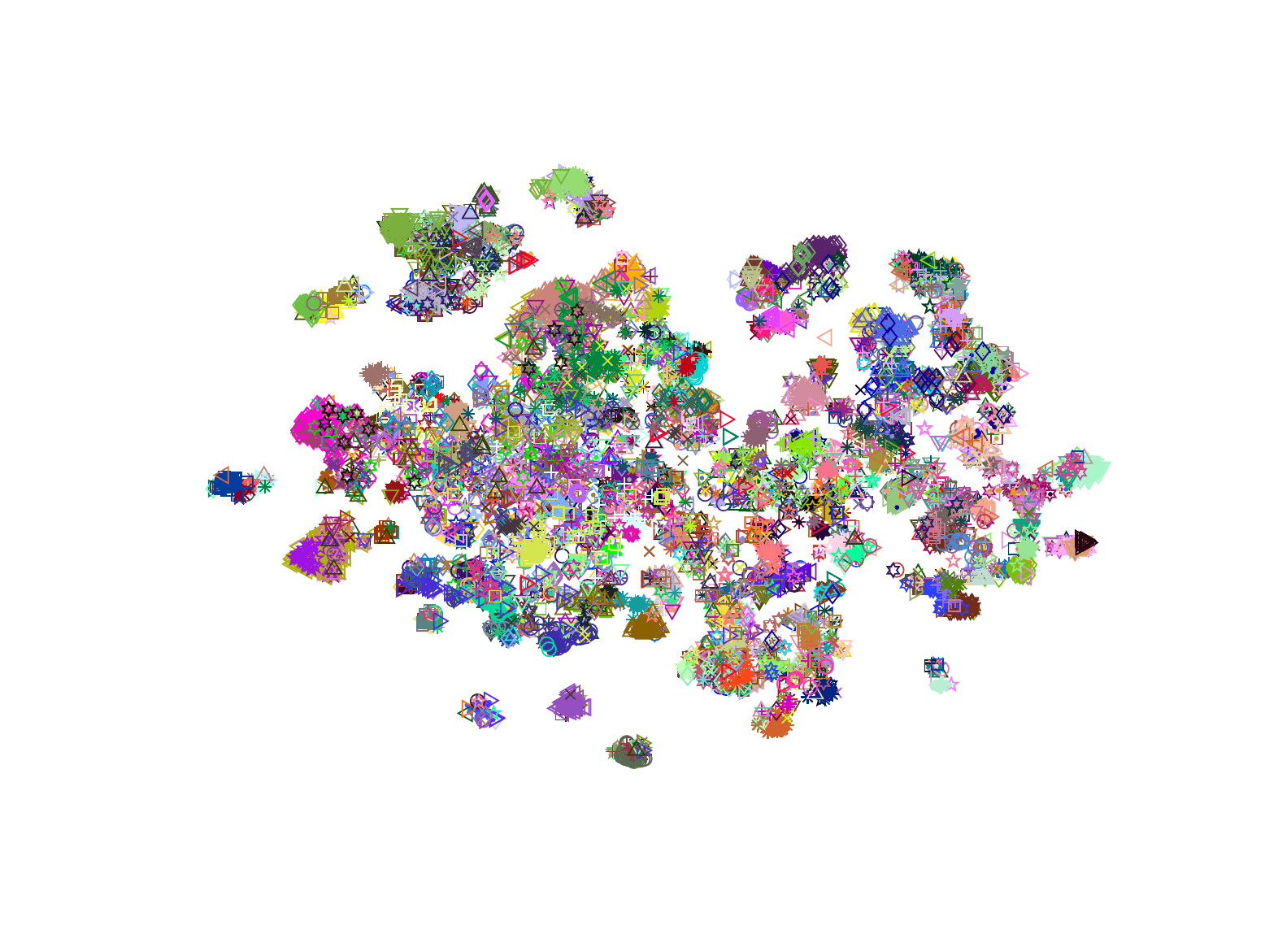}
        \caption{Width Augmented Network (WA-CNN)}
    \end{subfigure}%
  \end{center}
\vspace{-0.5cm}
   \caption{t-SNE visualizations of the top feature layers on the SUN-$397$ validation set. DA-CNN and WA-CNN show significantly better semantic separations.}
\label{fig:tsne}
\vspace{-.3cm}
\end{figure}

\textbf{Feature visualization:} To roughly understand the topology of the feature spaces, we visualize the features using the standard t-SNE algorithm~\cite{maaten2008visualizing}. As shown in Figure~\ref{fig:tsne}, we embed the \begin{small}$4{,}096$\end{small}-dim \begin{small}$FC_7$\end{small} features of the pre-trained and fine-tuned networks, the \begin{small}$6{,}144$\end{small}-dim wider  \begin{small}$FC_7+FC_7^+$\end{small} features, and the \begin{small}$2{,}048$\end{small}-dim deeper \begin{small}$FC_a$\end{small} features into a \begin{small}$2$\end{small}-dim space, respectively, and plot them as points colored depending on their
semantic category. While classic fine-tuning somehow improves the semantic separation of the pre-trained network, both of our networks demonstrate significantly clearer semantic clustering structures, which is compatible with their improved classification performance.

\begin{figure*}[t]
\begin{center}
\centering
   \includegraphics[trim=1.5cm 1.6cm 1.5cm 0.5cm,clip=true,width=\linewidth]{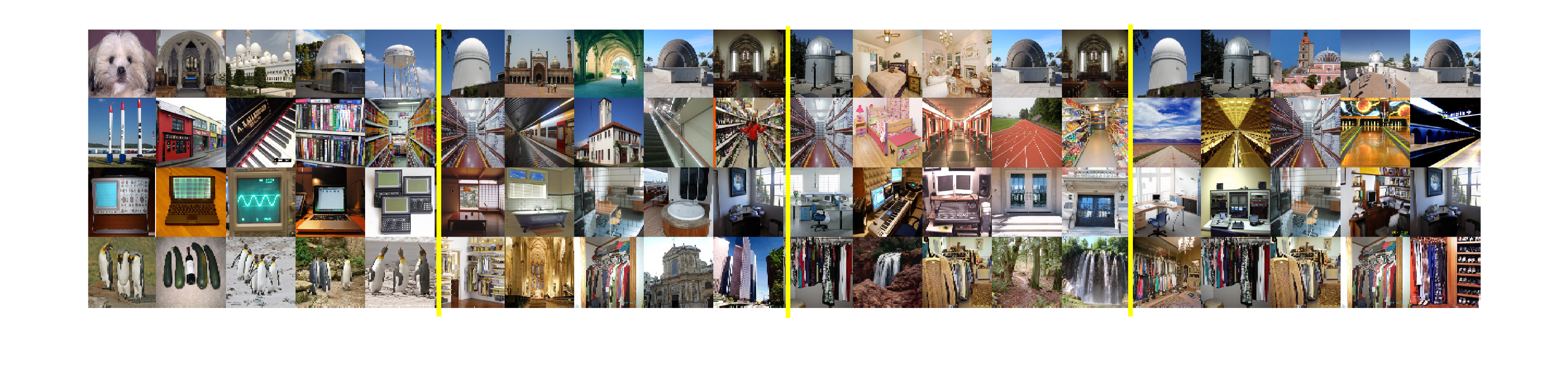}
\end{center}
\vspace{-0.5cm}
   \caption{Top $5$ maximally activating images for four $FC_7$ units. From left to right: ILSVRC $2012$ validation images for the pre-trained network, and SUN-$397$ validation images for the pre-trained, fine-tuned, and width augmented (WA-CNN) networks. Each row of images corresponds to a common unit from these networks, indicating that our WA-CNN facilitates the specialization of the pre-existing units towards the novel target task. For example, the bottom row shows a transition from a penguin-like vertically repeated pattern in the pre-trained ImageNet network to several mixed concepts in the fine-tuned network, and finally to a wardrobe-like vertically repeated pattern in our SUN-$397$ WA-CNN.}
\label{fig:maxwider}
\vspace{-0.3cm}
\end{figure*}

\begin{figure}[t]
\begin{center}
\centering
   \includegraphics[trim=0cm 7cm 0cm 7cm,clip=true,width=.48\linewidth]{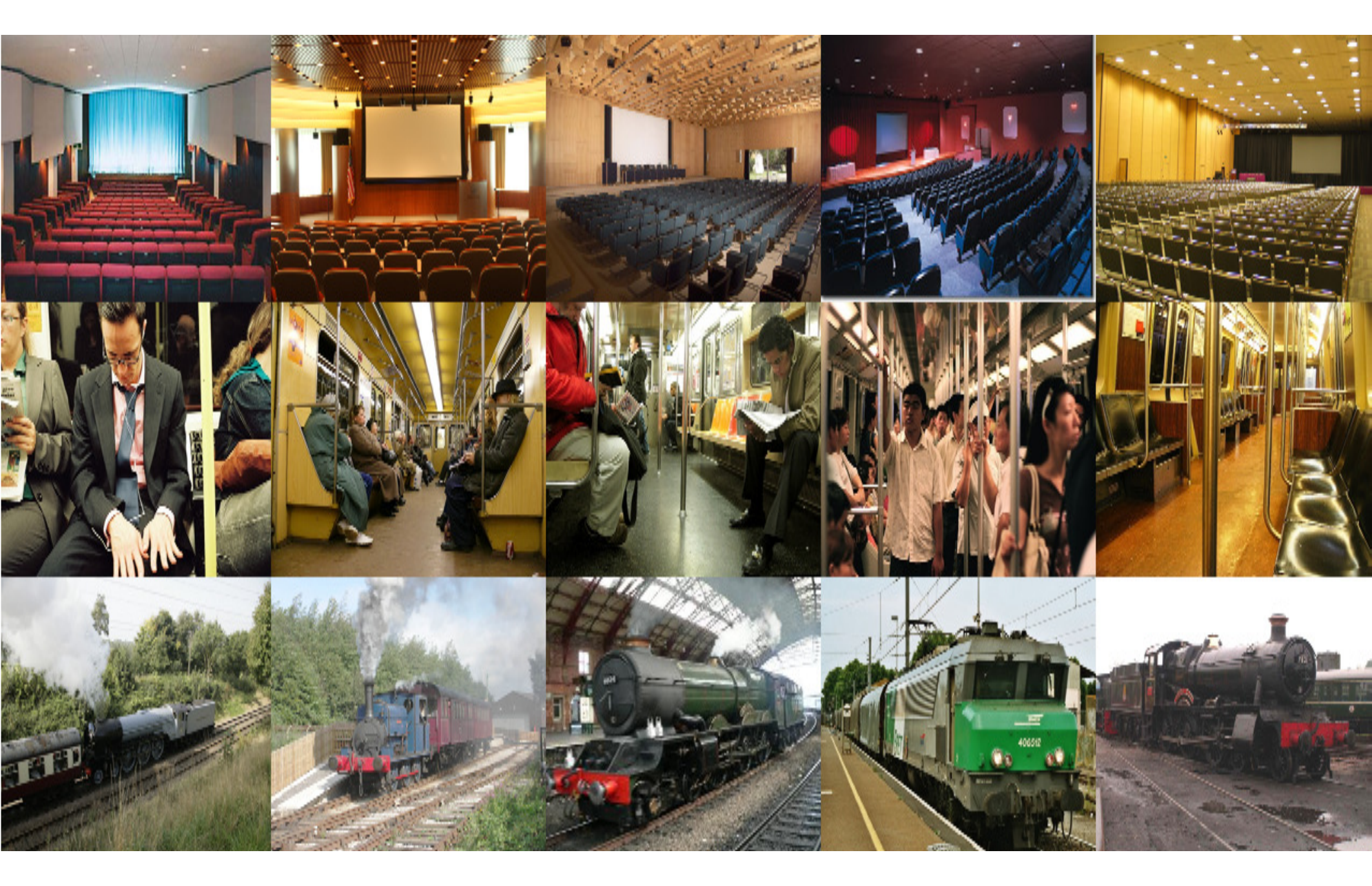}
~
   \includegraphics[trim=0cm 7cm 0cm 7cm,clip=true,width=.48\linewidth]{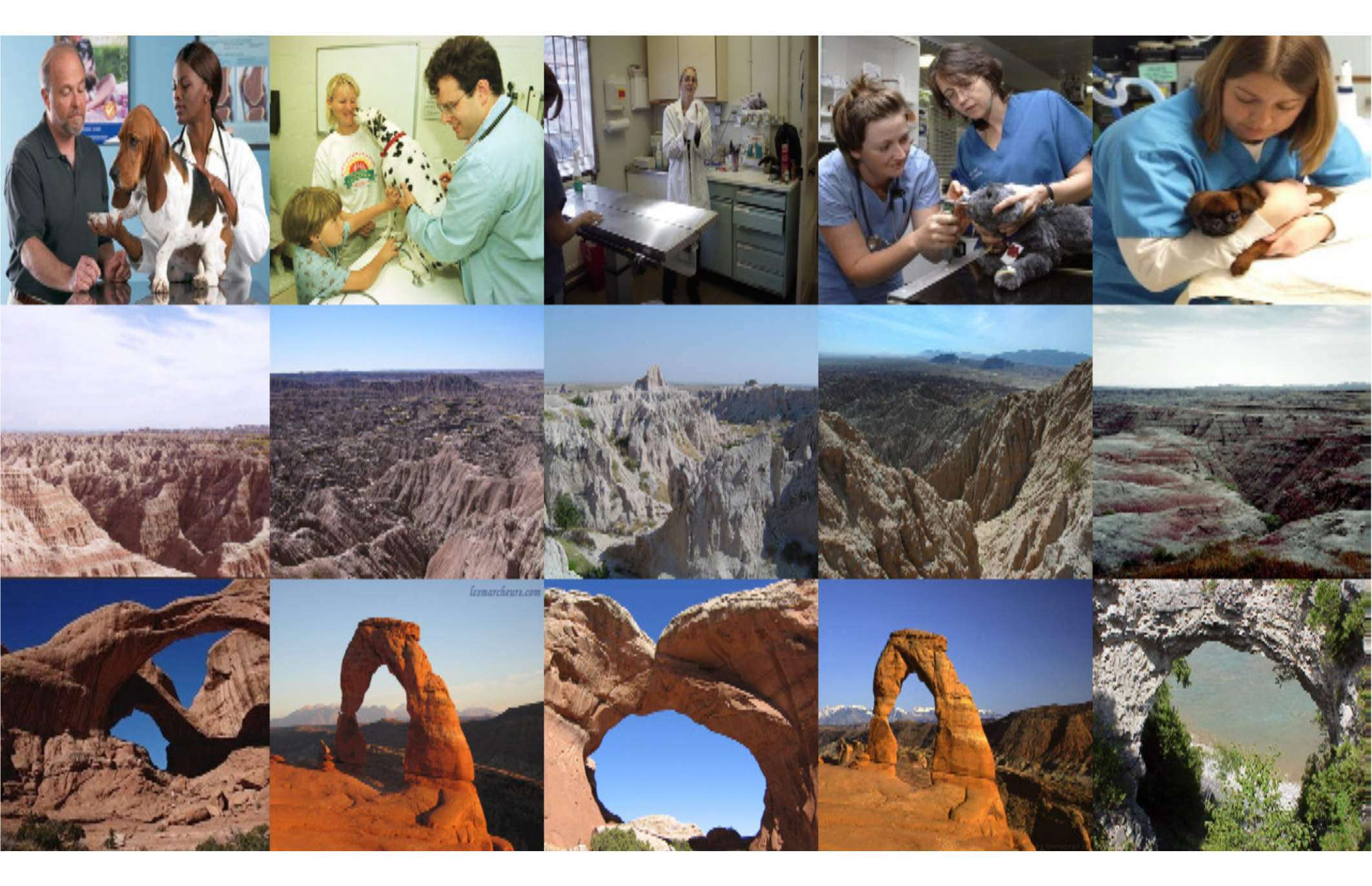}
\end{center}
\vspace{-0.4cm}
   \caption{Top $5$ maximally activating images from the SUN-$397$ validation set for six $FC_a$ units of the depth augmented network (DA-CNN). Each row of $5$ images in the left and right columns corresponds to a unit, respectively, which is well aligned to a scene-level concept for the target task, \eg, auditorium and veterinary room in the first row.} 
\label{fig:maxdeeper}
\vspace{-0.3cm}
\end{figure}

\textbf{Maximally activating images:} To further analyze how fine-tuning changes the feature spaces, we retrieve the top-$5$ images that maximally activate some unit as in~\cite{girshick2014rich}. We first focus on the {\em common units} in \begin{small}$FC_7$\end{small} of the pre-trained, fine-tuned, and width augmented networks. In addition to using the SUN-$397$ images, we also include the maximally activating images from the ILSVRC $2012$ validation set for the pre-trained network as references. Figure~\ref{fig:maxwider} shows an interesting transition: while the pre-trained network learns certain concentrated concept specific to the source task (left), such concept spreads over as a mixture of concepts for the novel target task (middle left). Fine-tuning tries to re-centralize one of the concepts suitable to the target task, but with limited capability (middle right). Our width augmented network facilitates such re-centralization, leading to more discriminative patterns (right). Similarly, we illustrate the maximally activating images for units in \begin{small}$FC_a$\end{small} of the depth augmented network in Figure~\ref{fig:maxdeeper}, which shows quite different behaviors. Compared with the object-level concepts in the width augmented network, the depth augmented network appears to have the ability to model a large set of compositions of the pre-trained features and thus generates more scene-level, better clustered concepts.

\begin {table*}[h]
\centering
\begin{minipage}{\textwidth}
\centering
\resizebox{\linewidth}{!}{
\begin{tabular}{l|l| c|l|c|l|c|l|c }
\hline
  \multirow{2}{*}{Type}& \multicolumn{2}{c|}{MIT-67} & \multicolumn{2}{c|}{102 Flowers} &\multicolumn{2}{c|}{CUB200-2011}& \multicolumn{2}{c}{Stanford-40}\\ \cline{2-9}
  & Approach & Acc(\%) & Approach & Acc(\%) & Approach & Acc(\%) & Approach & Acc(\%)  \\ \hline 
  \multirow{3}{*}{ImageNet CNNs}
  & Finetuning-CNN & 61.2 & Finetuning-CNN & 75.3 & Finetuning-CNN & 62.9& Finetuning-CNN & 57.7\\
  &Caffe~\cite{yang2015multi} & 59.5 &CNN-SVM~\cite{razavian2014cnn} & 74.7 &CNN-SVM~\cite{razavian2014cnn} & 53.3&Deep Standard~\cite{azizpour2015generic} & 58.9\\ 
  & ---& ---&CNNaug-SVM~\cite{razavian2014cnn} & 86.8 &CNNaug-SVM~\cite{razavian2014cnn} & 61.8 & ---&---\\ \hline 
  \multirow{4}{*}{\tabincell{c}{Task Customized\\CNNs}}  
  &Caffe-DAG~\cite{yang2015multi} & 64.6 &LSVM~\cite{qian2015fine} & 87.1 &LSVM~\cite{qian2015fine} & 61.4&Deep Optimized~\cite{azizpour2015generic} & 66.4\\ 
  & --- & --- &MsML+~\cite{qian2015fine} &  89.5 &DeCaf+DPD~\cite{donahue2014decaf}&65.0& ---&---\\
  &Places-CNN~\cite{zhou2014learning} & \bf{68.2} &MPP~\cite{yoo2015multi} & 91.3 &MsML+~\cite{qian2015fine} & 66.6 &--- &---\\ 
  &--- & --- &Deep Optimized~\cite{azizpour2015generic} & 91.3 & MsML+*~\cite{qian2015fine} & 67.9& ---&---\\\hline
  \multirow{1}{*}{\tabincell{c}{Data Augmented CNNs}}
  &Combined-AlexNet~\cite{joulin2016learning} & 58.8 &Combined-AlexNet~\cite{joulin2016learning} & 83.3 & ---& ---&Combined-AlexNet~\cite{joulin2016learning} & 56.4\\\hline
  \multirow{2}{*}{\tabincell{c}{Multi-Task CNNs}}
  & Joint~\cite{li2016learning} & 63.9 & ---& ---&Joint~\cite{li2016learning} & 56.6 &--- &---\\ 
  &LwF~\cite{li2016learning} & 64.5 &--- &--- &LwF~\cite{li2016learning} & 57.7 &--- &---\\ 
  \hline\hline
  \multirow{1}{*}{Ours}
  &WA-CNN & 66.3&WA-CNN & \textbf{92.8}&WA-CNN & \bf{69.0}&WA-CNN  & \bf{67.5}\\
  \hline
\end{tabular}
}
\label{tab:mit67}
\end{minipage}
\vspace{-0.2cm}
\caption {Performance comparisons of classification accuracy (\%) between our developmental networks (WA-CNN) and the previous work for scene classification, fine-grained recognition, and action recognition. We roughly divide the baselines into four types: (1) ImageNet CNNs, which post-process the off-the-shelf CNN or fine-tune it in a standard manner; (2) task customized CNNs, which modify a standard CNN for a particular target task (\eg, for MIT-$67$, Places-CNN trains a customized CNN on the Places dataset with $400$ scene categories~\cite{zhou2014learning}); (3) data augmented CNNs, which concatenate features from the ImageNet AlexNet and an additional CNN trained on $100$ million Flickr images in a weakly supervised manner~\cite{joulin2016learning}; (4) multi-task CNNs, which (approximately) train a CNN jointly from both the source and target tasks. Ours show consistently superior performance and generality for a wide spectrum of tasks.}
\label{tab:fullset}
\vspace{-0.5cm}
\end {table*}

\subsection{Generalization to Other Tasks and Datasets}
\label{sec:novelexp}

We now evaluate whether our developmental networks facilitate the recognition of other novel categories. We compare with publicly available baselines and report multi-class classification accuracy. While the different variations of our networks outperform these baselines, we mainly focus on the width augmented networks (WA-CNN).

\textbf{Tasks and datasets:} We evaluate on standard benchmark datasets for scene classification: MIT-$67$~\cite{quattoni2009recognizing}, for fine-grained recognition: Caltech-UCSD Birds (CUB) $200$-$2011$~\cite{wah2011caltech} and Oxford $102$ Flowers~\cite{nilsback2008automated}, and for action recognition: Stanford-40 actions~\cite{yao2011human}. These datasets are widely used for evaluating the CNN transferability~\cite{azizpour2015factors}, and we consider their diversity and coverage of novel categories. We follow the standard experimental setup (\eg, the train/test splits) for these datasets.

\textbf{Baselines:} While comparing with classic fine-tuning is the fairest comparison, to show the superiority of our approach, we also compare against other baselines that are specifically designed for certain tasks. For a fair comparison, we focus on the approaches that use single scale AlexNet CNNs. Importantly, our approach can be also combined with other CNN variations (\eg, VGG-CNN~\cite{simonyan2015very}, multi-scale CNN~\cite{hariharan2015hypercolumns,yang2015multi}) for further improvement.

Table~\ref{tab:fullset} shows that our approach achieves state-of-the-art performance on these challenging benchmark datasets and significantly outperforms classic fine-tuning by a large margin. In contrast to task customized CNNs that are only suitable for particular tasks and categories, the consistently superior performance of our approach suggests that it is generic for a wide spectrum of tasks.

\vspace{-0.1cm}
\subsection{A Single Universal Higher Capacity Model?}
\label{sec:singlemodel}
\vspace{-0.1cm}
One interesting question is that our results could imply that standard models should have used higher capacity even for the source task (\eg, ImageNet). To examine this, we explore progressive widening of AlexNet (WA-CNN). Specifically, in the source domain, Table~\ref{tab:single-scratch} shows that progressive widening of a network outperforms a fixed wide network trained from scratch. More importantly, in the target domain, Table~\ref{tab:single-augorder} shows that our progressive widening {\em significantly outperforms} fine-tuning a fixed wide network.

\begin{table}[t]
\vspace{-0.5cm}
\small
\begin{center}
\begin{tabular}{c| c|c|c}
\hline
Dataset & CNN & WA-CNN-scratch &WA-CNN-grow (Ours)\\ \hline
ImageNet & 56.9 & 57.6 & \bf{57.8}\\\hline
\end{tabular}
\vspace{-0.2cm}
\caption{Performance comparisons of classification accuracy (\%) on the source dataset between a standard AlexNet (CNN), a wide AlexNet trained from scratch (WA-CNN-scratch), and a wide network trained progressively by fine-tuning on the source task itself (WA-CNN-grow). Progressive learning appears to help even on the source task.}

\label{tab:single-scratch}
\end{center}
\vspace{-0.7cm}
\end{table}

\begin{table}[t]
{\small
\begin{center}
\begin{tabular}{c| c|c|c}
\hline
Dataset & CNN-FT & WA-CNN-ori &WA-CNN-grow (Ours)\\ \hline
MIT-67 & 61.2 & 62.3 & \bf{66.3}\\
\hline
CUB200-2011 & 62.9 & 63.2 & \bf{69.0} \\ \hline
\end{tabular}
\vspace{-0.2cm}
\caption{Performance comparisons of classification accuracy (\%) on the target datasets between standard fine-tuning of a standard AlexNet (CNN-FT), standard fine-tuning of a wide AlexNet (WA-CNN-ori), and fine-tuning by progressive widening of a standard AlexNet (WA-CNN-grow).
With the same model capacity, WA-CNN-grow significantly outperforms WA-CNN-ori. See Figure~\ref{fig:single-learning} for a discussion.}
\label{tab:single-augorder}
\end{center}
\vspace{-0.69cm}
}
\end{table}

\begin{figure}[h]
\begin{center}
\centering
   \includegraphics[trim=1.2cm 0.2cm 3.8cm 0.4cm,clip=true,width=.45\linewidth]{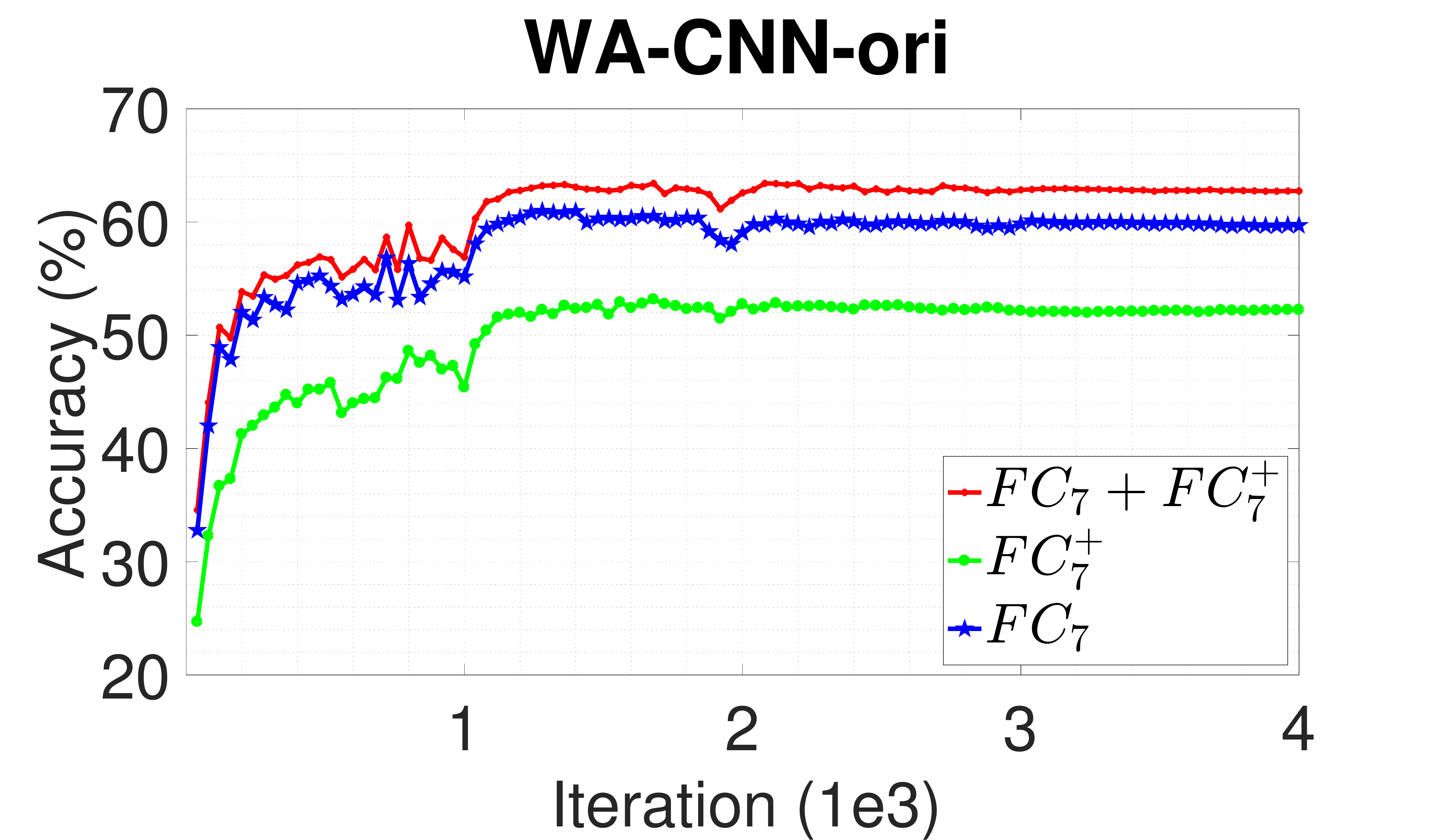}
   ~
      \includegraphics[trim=1.2cm 0.2cm 3.8cm 0.4cm,clip=true,width=.45\linewidth]{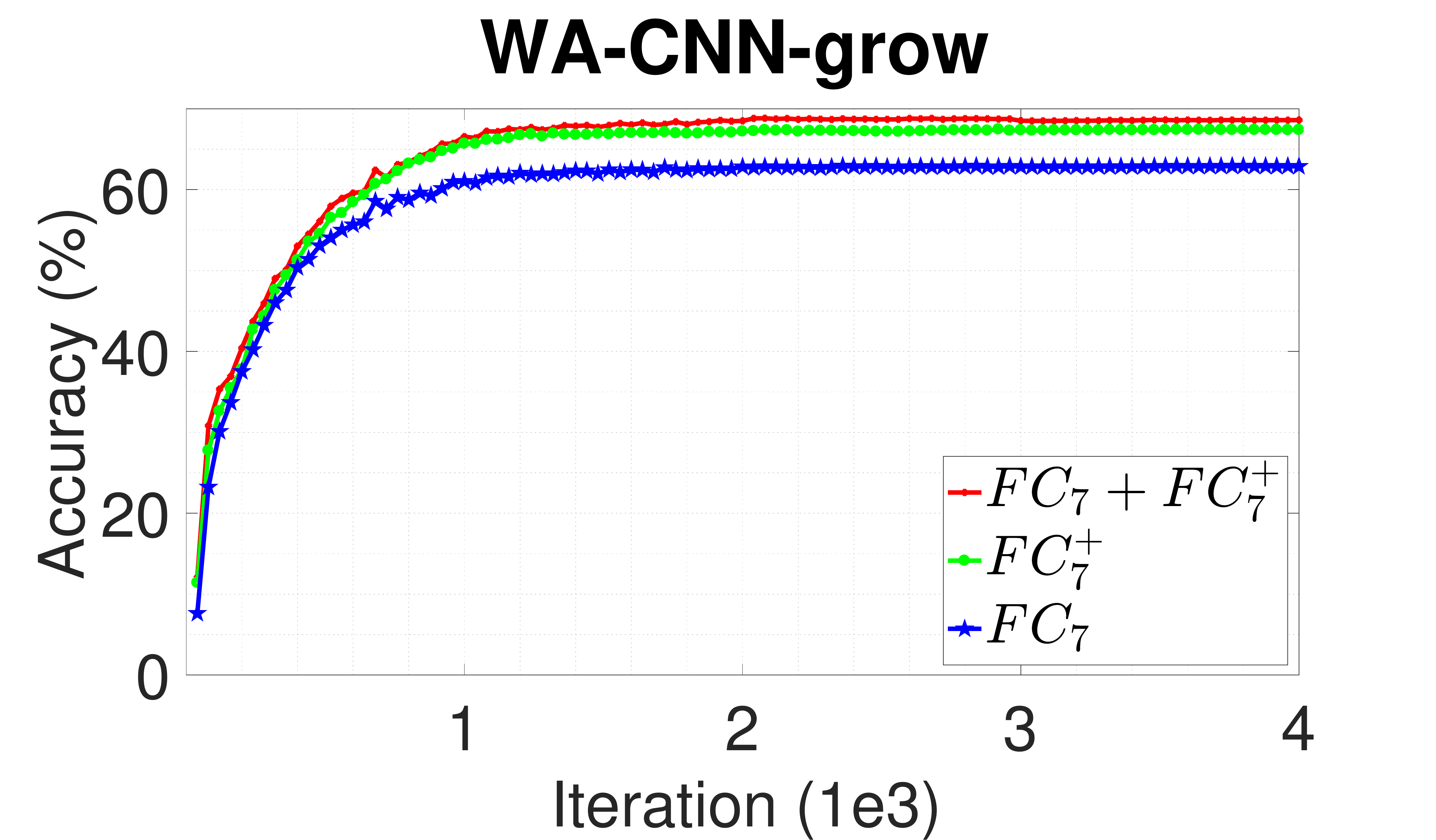}
  \end{center}
\vspace{-0.5cm}
   \caption{Learning curves of separate $FC_7$ and $FC_7^+$ and their combination for WA-CNN on the CUB$200$-$2011$ test set. Left and Right show different learning behaviors: the $FC_7^+$ curve is {\em below} the $FC_7$ curve for WA-CNN-ori, and {\em above} for WA-CNN-grow. Units in WA-CNN-ori appear to {\em overly-specialize} to the source, while the new units in WA-CNN-grow appear to be {\em diverse experts} better tuned for the novel target task. Interestingly, these experts allow for better adaptation of pre-existing and new units (Figure~\ref{fig:single-maxwider}).}
\label{fig:single-learning}
\end{figure}

\begin{figure}[h]
\begin{center}
\centering
   \includegraphics[trim=2.3cm 7.8cm 2.3cm 1cm,clip=true,width=\linewidth]{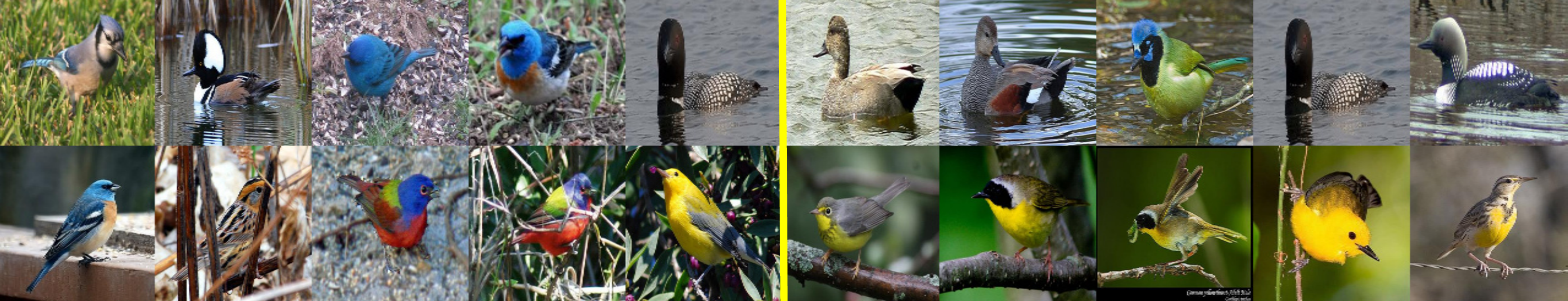}
\end{center}
\vspace{-0.5cm}
   \caption{Top $5$ maximally activating CUB$200$-$2011$ images for a representative $FC_7$ unit ($1$st row) and an $FC_7^+$ unit ($2$nd row). Each row of images corresponds to a common unit from two networks: WA-CNN-ori (left) and WA-CNN-grow (right). Compared to WA-CNN-ori, WA-CNN-grow facilitates the adaptation of pre-existing and new units towards the novel task by capturing discriminative patterns (top: birds in water; bottom: birds with yellow belly).}
\label{fig:single-maxwider}
\end{figure}

\textbf{Cooperative learning:} 
Figure~\ref{fig:single-learning} and Figure~\ref{fig:single-maxwider} provide an in-depth analysis of the cooperative learning behavior between the pre-existing and new units and show that developmental learning appears to regularize networks in a manner that encourages diversity of units.

\textbf{Continual transfer across multiple tasks:} Our approach is in particular suitable for {\em continual, smooth} transfer across multiple tasks since we are able to cumulatively increase model capacity as demonstrated in Table~\ref{tab:multi}.
\begin{table}[h]
\small
\begin{center}
\resizebox{\linewidth}{!}{
\begin{tabular}{c| c|c|c|c}
\hline
\multirow{2}{*}{Scenario}& \multicolumn{2}{c|}{WA-CNN (Ours)} &\multicolumn{2}{c}{Baselines} \\ \cline{2-5}
& ImageNet$\rightarrow$MIT67 & ImageNet$\rightarrow$SUN$\rightarrow$MIT67 & Places~\cite{zhou2014learning} & ImageNet-VGG~\cite{li2016learning} \\ \hline
Acc(\%)& 66.3 & \bf{79.3} & 68.2 & 74.0\\ \hline	
\end{tabular}
}
\vspace{-0.2cm}
\caption{Through progressive growing via SUN-$397$, a widened AlexNet significantly improves the performance on MIT-$67$, and even outperforms fine-tuning a Places AlexNet that is directly trained on the Places dataset with $400$ scene categories~\cite{zhou2014learning} and fine-tuning a fixed ImageNet VGG16 with higher capacity by a large margin.}
\label{tab:multi}
\end{center}
\vspace{-0.8cm}
\end{table}

%---------------------------------------------------------------
\section{Conclusions}
We have performed an in-depth study of the ubiquitous practice of fine-tuning CNNs. By analyzing what changes in a network and how, we conclude that increasing model capacity significantly helps existing units better adapt and specialize to the target task. We analyze both depth and width augmented networks, and conclude  that they are useful for fine-tuning, with a slight but consistent benefit for widening. A practical issue is that newly added units should have a pace of learning that is comparable to the pre-existing units. We provide a normalization and scaling technique that ensures this. Finally, we present several state-of-the-art results on benchmark datasets that show the benefit of increasing model capacity. Our conclusions support a developmental view of CNN optimization, in which model capacity is progressively grown throughout a lifelong learning process when learning from continuously evolving data streams and tasks.

 {\bf{Acknowledgments:}} We thank Liangyan Gui for valuable and insightful discussions. This work was supported in part by ONR MURI N000141612007 and U.S. Army Research Laboratory (ARL) under the Collaborative Technology Alliance Program, Cooperative Agreement W911NF-10-2-0016. DR was supported by NSF Grant
1618903 and Google. We also thank NVIDIA for donating GPUs and AWS Cloud Credits for Research program.

\clearpage
{\small
\bibliographystyle{ieee}
\bibliography{egbib}
}

\end{document}